\documentclass[conference]{IEEEtran}
\IEEEoverridecommandlockouts
\usepackage{cite}
\usepackage{amsmath,amssymb,amsfonts}
\usepackage{algorithmic}
\usepackage{graphicx}
\usepackage{textcomp}

\makeatletter

\renewcommand\section{\@startsection {section}{1}{\z@}%
                                   {-1.2ex \@plus -2ex \@minus -.2ex}%
                                   {.5ex \@plus.2ex}%
                                   {\normalfont\large\bfseries}}
\renewcommand\subsection{\@startsection{subsection}{2}{\z@}%
                                   {-.6ex \@plus -2ex \@minus -.2ex}%
                                   {.5ex \@plus.2ex}%
	                           {\normalfont\large\bfseries}}
\renewcommand\subsubsection{\@startsection{subsubsection}{3}{\z@}%
                                     {-.5ex\@plus -.2ex \@minus -.2ex}%
                                     {.2ex \@plus .2ex}%
                                     {\normalfont\large\bfseries}}

\let\paragraph\textbf
\makeatother

\def\tightmath{
\abovedisplayskip=4pt plus 2pt minus 1pt 
\abovedisplayshortskip=2pt plus 1pt minus 1pt 
\belowdisplayskip=4pt plus 2pt minus 1pt 
\belowdisplayshortskip=2pt plus 1pt minus 1pt }

\tightmath

\usepackage{multirow}
\usepackage{xcolor,colortbl}
\usepackage{float}
\definecolor{darkergreen}{RGB}{21, 152, 56}
\definecolor{red2}{RGB}{252, 54, 65}
\definecolor{greencol}{RGB}{0, 255, 0}
\definecolor{redcol}{RGB}{252, 0, 0}
\definecolor{bluecol}{RGB}{0, 0, 255}

\definecolor{mygray}{gray}{.88}
\definecolor{mygraylite}{gray}{.93}

\usepackage{pifont}% http://ctan.org/pkg/pifont
\newcommand{\xmark}{\ding{55}}%

\newcommand\ie{i.e.}
\newcommand\eg{e.g.}

\usepackage{xcolor}
\usepackage[pagebackref=true,breaklinks=true,colorlinks,bookmarks=false]{hyperref}
\begin{document}
%\captionsetup[figure]{labelfont={small,bf},textfont={small,it},name={Fig}}
%\captionsetup[table]{labelfont={small,bf},textfont={small,it},name={Tab}}

\title{Enhanced Performance of Pre-Trained Networks by Matched Augmentation Distributions}

\makeatletter
\newcommand{\linebreakand}{%
  \end{@IEEEauthorhalign}
  \hfill\mbox{}\par
  \mbox{}\hfill\begin{@IEEEauthorhalign}
}
\makeatother

%\author{\IEEEauthorblockN{Anonymous Authors}
%\IEEEauthorblockA{Anonymous Authors}
%{\tt\small \{Anonymous Authors\}}}

\author{\IEEEauthorblockN{Touqeer Ahmad, Mohsen Jafarzadeh, Akshay Raj Dhamija, Ryan Rabinowitz,\\ Steve Cruz, Chunchun Li, Terrance E. Boult}
\IEEEauthorblockA{Vision and Security Technology Lab, University of Colorado at Colorado Springs, USA}
{\tt\small \{touqeer, mjafarzadeh, adhamija, rrabinow, cli, scruz, tboult\}@vast.uccs.edu}}

\maketitle

\begin{abstract}
There exists a distribution discrepancy between training and testing, in the way images are fed to modern CNNs. Recent work tried to bridge this gap either by fine-tuning or re-training the network at different resolutions. However re-training a network is rarely cheap and not always viable. To this end, we propose a simple solution to address the train-test distributional shift and enhance the performance of pre-trained models -- which commonly ship as a package with deep learning platforms \eg, PyTorch. Specifically, we demonstrate that running inference on the center crop of an image is not always the best as important discriminatory information may be cropped-off. Instead we propose to combine results for multiple random crops for a test image. This not only matches the train time augmentation but also provides the full coverage of the input image. We explore combining representation of random crops through averaging at different levels \ie, deep feature level, logit level, and softmax level. We demonstrate that, for various families of modern deep networks, such averaging results in better validation accuracy compared to using a single central crop per image. The softmax averaging results in the best performance for various pre-trained networks without requiring any re-training or fine-tuning whatsoever. On modern GPUs with batch processing, the paper's approach to inference of pre-trained networks, is essentially free as all images in a batch can all be processed at once.  
\end{abstract}

\begin{figure}[t!]
\centering
\includegraphics[width=.75\linewidth]{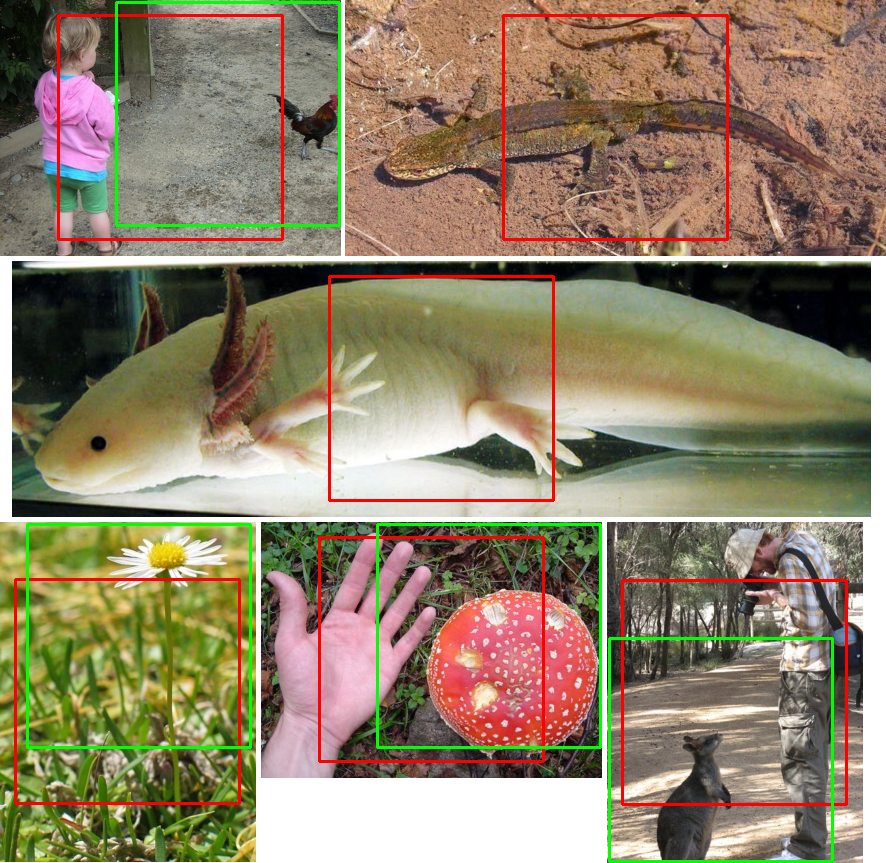}
\caption{Examples from ILSVRC-2012 \cite{Deng2009ImageNet} val split where a center-crop (\textcolor{redcol}{red} square) may miss discriminatory information e.g., rows one \& two, or at best provides partial coverage of object of interest \eg, row three (daisy, mushroom, kangaroo). In some of such cases, a non-central random crop (\textcolor{greencol}{green} square) may provide good coverage of the image and that of underlying object of interest. Whereas in other cases a single random or central crop may not be optimal and an average of several random crops may serve best. Each image above is resized to $256$ respecting aspect ratio and central $224\times224$ crop identified by red square which is the conventional input to standard networks (e.g., ResNet-50) at inference time.}
\label{fig_teaser}
\end{figure}

\section{Introduction}

Boosting the performance of established deep networks is an active research area where methods including custom training \cite{Touvron2019Fixing}, employing additional training sets \cite{Yalniz2019Billion-scale}, ensemble teacher-student paradigms \cite{Shen2021MEAL, Shen2019MEAL}, architecture modifications, complex learning schedules and data augmentation \cite{Yun2019CutMix, Zhang2018mixup, DeVries2017Improved, Zhong2020Random} strategies have been investigated. However re-training or fine-tuning an existing model with custom settings is not always a possible or preferred solution.

This paper's approach improves performance using a pre-trained network by addressing the discrepancies that exist between training and inference of deep neural networks. For example, during training data augmentation with random crops are generated from the images on which the loss is minimized, whereas, during inference it is conventional to resize an image to a fixed resolution maintaining aspect ratio and then take the central crop to forward pass it through the trained network. In most cases the central crop provides a reasonable coverage of the image and underlying object of interest. However in some cases centrally cropping an image may discard the discriminatory information essential for a good recognition. Fig \ref{fig_teaser} shows several examples from ILSVRC-2012 \cite{Deng2009ImageNet} validation set where a central crop may result into insufficient coverage of the resized image and especially of the object of interest.  

One can easily identify two types of categories of images which are at disadvantage due to central cropping: (i) images where the captured object of interest is not in the center of the frame, and (ii) images where longer dimension is significantly larger than the shorter one (more than 2x). In first category, since the object of interest is already at a non-central location, cropping centrally further removes or reduces the discriminatory information essential for the recognition. For example in Fig \ref{fig_teaser}, non-centrally captured objects of interest [row3: mushroom, kangaroo] are either partially covered by the central crop or fully cropped off in more severe cases [row1: roaster, row3: daisy]. In second category, the object of interest is generally laid out along the longer dimension of the image, central copping only captures part of the object and may discard important discriminatory information. For example the heads and tails of the animals imaged in row1 (right) and row2 of Fig \ref{fig_teaser} are discarded as a result of center cropping and characteristics of these parts could be the discriminatory features that the deep network learned during training.

To enhance the coverage of the underlying object of interest, some networks used multiple fixed crops at inference time \cite{Some2014Howard, Szegedy2015Going}, \eg, the original AlexNet \cite{krizhevsky2012imagenet}, used 5 fixed crops (center and 4 corners), plus their horizontal mirrored version for 10 crops at inference. While AlexNet did use multiple crops at inference, it still has a significant difference between training, which has 2048 random variations, and test time which has 5 fixed crop locations. Through an ablation study we demonstrate that while AlexNet's 5/10 crop strategy may be sufficient for most images, it can still benefit when more random or mirrored random crops are additionally added.     

Recent work such as FixRes \cite{Touvron2019Fixing}, show that a performance gain can be obtained by adjusting resolution so the distribution of object sizes is better matched between training and inference. They accomplish this size-distribution gap by retraining the network. In this paper we content that the size mismatch is only part of the distributional mismatch problem, and that object coverage via crops is also an issue. To remedy the recognition performance, we propose Matched Inference Distributions (MID) which approximates the distribution of training sampling at inference time, and show this improves performance without the need for any retraining.  In training, over the multiple epochs the system effectively averages the different sample augmentations. Hence MID combines results by averaging from several augmentations instead of forward passing a single central crop per image. 

The benefit of using random crops is three fold: (i) it better aligns with augmentation employed during the training of deep networks, (ii) provides good coverage of underlying object of interest, and (iii) it addresses some of the resolution issues identified in FixRes \cite{Touvron2019Fixing}.  The representations generated for several random crops of an image can be combined at different levels of a deep network. To this end we explore averaging at the levels of deep features, logits and the softmax layer. We found that averaging after deep features and logits results in identical performance whereas averaging at softmax layers results in the best performance. We have investigated our averaging of random crops approach for various modern network families including ResNet \cite{He2016Deep}, EfficientNet \cite{Tan2019EfficientNet}, and NFNet \cite{Brock2021High} with up to \textbf{2.0\%} boost in ImageNet-2012 Top-1 validation accuracy. We should note that the performance gain is achieved without requiring any custom training, re-training or fine-tuning of the networks. Since methods focusing on better training strategies \cite{Shen2019MEAL,Shen2021MEAL,Yalniz2019Billion-scale,Touvron2019Fixing} still run inference using a single central crop, they can further benefit from proposed averaging strategy. We should note that batch processing is common in modern deep learning platforms and forward passing multiple crops can by achieved without any significant overhead. For a non-batched processing, forward passing multiple crops directly results in enhanced computations. To address this we suggest an adaptive strategy where the number of random crops are pre-determined as function of input resolution as center cropping may still be sufficient for majority of the test images.

\paragraph{Our Contributions}
\begin{itemize}
\item Demonstrating that central or simple fixed crops at inference are sub-optimal and that our novel Matched Inference Distribution (MID) approach which uses the same sampling process for data augmentation in both training and inference, results in better performance.   
\item Thorough evaluation of several families of modern deep networks to quantify their achievable best performance without any custom training or fine-tuning.
%\item An adaptive strategy to choose the number of random crops in accordance to the input resolution.  
\end{itemize}

\begin{figure*}[t!]
\centering
\includegraphics[width=.7\linewidth]{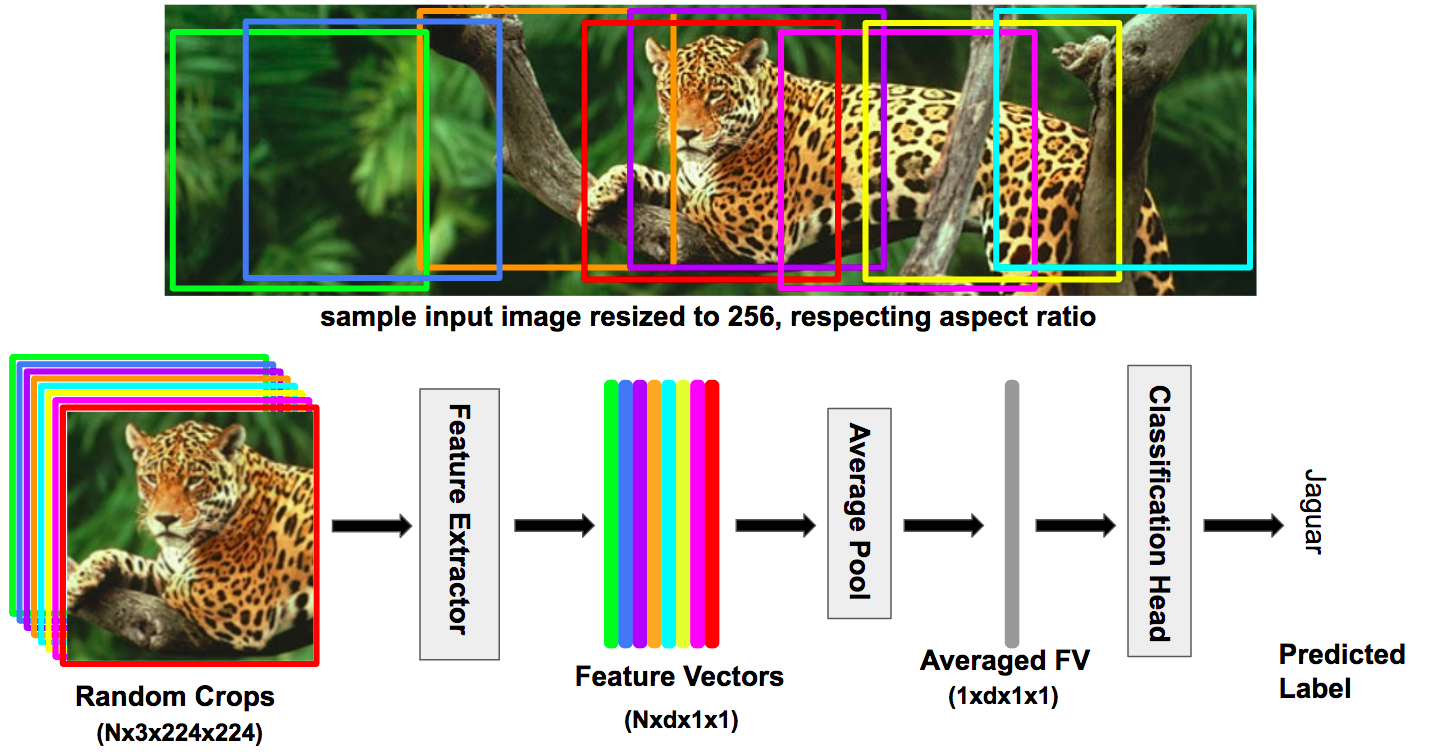}
\caption{One variant of our proposed inference scheme where we advocate using averaging of several random crops instead of just a single central one. A sample image from ILSVRC-2012 \cite{Deng2009ImageNet} validation split, resized to 256 respecting aspect ratio with several $224 \times 224$ random crops marked in different colors. Adopting PyTorch's preferred NCHW tensor format, the generated random crops are concatenated in the batch (N) dimension and passed through the feature extractor of a pre-trained CNN. The generated deep feature vectors are then averaged and subsequently passed through the classification head to predict the classification label. Softmax layer is omitted for brevity.}
\label{fig_approach}
\end{figure*}

%-------------------------------------------------------------------------
\section{Related Work}
\label{sec:Related}
Squeezing the best performance out of established network architectures for image recognition is an active research area with supreme practical importance of real-world deployment especially of smaller networks like ResNet-18, and MobileNet etc. Latest attempts have been focused on training, re-training or fine-tuning the core networks with custom strategies using default or additional data.

In popular FixRes \cite{Touvron2019Fixing}, authors emphasized the existence of train-test object size discrepancy and argued for using smaller resolutions during training. Through extensive experiments they demonstrated that training on smaller resolutions not only compensates for the distributional shift but also reduces the training time. They further showed that fine-tuning pre-trained networks for higher test-time resolution results in enhanced performance. Using ResNet-50 as the core architecture they explored various train-test resolution combinations and demonstrated that even higher performance can be achieved by employing bigger networks like ResNeXt-101 \cite{Xie2017Aggregated} and leveraging Billion-scale training data \cite{Yalniz2019Billion-scale}. In an extended version \cite{Touvron2020Fixing} of their work, authors explored FixRes to train EfficientNet architectures \cite{Tan2019EfficientNet} where they further integrated label smoothing. They focused on two best performing EfficientNet versions trained with adversarial examples \cite{Xie2020Adversarial} and noisy student in weakly-supervised fashion using 300 million unlabeled images \cite{Xie2020Self-training}. The resultant FixEfficientNet models are demonstrated to outperform these two versions on ImageNet \cite{Deng2009ImageNet}, ImageNet-V2 \cite{Recht2019Do}, and ImageNet-Real \cite{Beyer2020Are}.            

CutMix \cite{Yun2019CutMix} belongs to the category of regional dropout augmentation methods where random regions in images are removed to enhance the generalization performance of deep networks. However unlike other regional dropout approaches \cite{DeVries2017Improved, Zhong2020Random}, CutMix fills the removed regions with patches from other training examples, effectively achieving not only better generalization but also better detection and localization performance. Mixup \cite{Zhang2018mixup} is another augmentation strategy where training images and ground truth labels are linearly interpolated to synthesize samples to enhance the generalization capability of the underlying network. Mixing features instead of the images \cite{Verma2019Manifold} and other mixup variants \cite{Summers2019Improved, Guo2019MixUp} have also been explored.               

In \cite{Shen2019MEAL}, authors explored knowledge distillation \cite{Hinton2015Distilling} and adversarial learning \cite{Goodfellow2014Generative} to train a student network using the predictions from an ensemble of teacher networks. Instead of using one-hot ground truth labels, the predicted probability vectors out of the teacher networks are used to distill the knowledge where L1, L2, and KL-divergence losses are explored. Additionally discriminator networks are incorporated at several levels to distinguish between the feature representations generated by student and the teacher. At each iteration a teacher is randomly selected out of the teacher zoo and its predicted probability vector provides the supervisory signal. For ImageNet, they used ResNet-50 as the student network and (i) VGG-19 w/BN and ResNet-50, or (ii) ResNet-101 and ResNet-152 as the two teacher ensemble variants. In extended version of their work \ie, MEAL-V2 \cite{Shen2021MEAL} they trained a student network using an ensemble of teachers where average of the softmax scores of the ensemble is used as soft supervisory signal instead of one-hot hard labels or selecting one teacher at random for every iteration. In this variant, the student network is trained using KL-divergence loss to match the probability distribution of the averaged ensemble. Like \cite{Shen2019MEAL}, a discriminator is further employed to distinguish whether the input features are generated from teacher ensemble or student network. Authors used their framework to train ResNet-50, EfficientNet-B0, and three size variants of MobileNet-V3 where they used senet154 and resnet152\_v1s as teachers. For training the student on a larger resolution \eg, $380 \times 380$, larger teachers like efficientnet\_b4\_ns and efficientnet\_b4 are employed.             

As evidenced by summarized research above, an enhanced performance can be achieved by smartly training the existing networks and/or leveraging the additional billion-scale data. However, training a network may not be a preferred or viable solution for vendors deploying deep models for real-world applications. To this end, in this paper we focus on squeezing best achievable performance for several popular pre-trained deep networks without any training or fine-tuning. We demonstrate that for several larger networks, averaging of random crops results in comparable or better performance than employing custom training strategies and billion-scale data rendering such strategies questionable.                  

%-------------------------------------------------------------------------
\section{Problem Formulation}
\label{sec:Formulation}

We operate in the conventional image recognition setting where we assume a pre-trained CNN model trained on train split of ILSVRC-2012 \cite{Deng2009ImageNet} is available. We further assume that the deep model can be decomposed into feature extractor $f_e$ and the classification head $f_c$. In the conventional single central-crop inference, a resized query image $x$ can be passed through the feature extractor $f_e$ to generate its deep representation:

\begin{equation}
R_{x^{c}} = f_e(x^{c})
\end{equation}

which can be subsequently passed through the classification head $f_c$ and the softmax layer to generate the predicted label:

\begin{equation}
\label{eq_prediction}
s_i = softmax(f_c(R_{x^{c}}))
\end{equation}

where, $x^{c}$ refers to the central crop generated from the query image $x$. As demonstrated in Fig \ref{fig_teaser}, a central-crop may not be ideal for all images, so we propose to use several random crops per image. Given the resized query image $x$, we generate $N$ random crops $x^{r_1}, x^{r_2}, \cdots, x^{r_N}$; each of which is passed through the feature extractor to generate respective representation $R_{x^{rj}}$. The representations for these random crops are then averaged:

\begin{equation}
R_{x^{avg}} = \frac{1}{N}\sum_{j=1}^{N}{R_{x^{rj}}}, 
\end{equation}

and passed through Eq. \ref{eq_prediction} to get the predicted label. Practically random crops can be concatenated to generate a batch which can be forward passed thought the feature extractor at once, and averaging can be accomplished by using the conventional AvgPool layer on the batch dimension. Fig \ref{fig_approach} provides a visualization for this formulation. Although the formulation has been described for averaging the representations at the deep feature level, similar averaging can be accomplished using the logits or softmax scores of random crops and have been explored in our experiments.

\begin{table*}[ht]
\centering
\caption{Comparison of \textbf{Top-1} validation accuracy on ImageNet dataset for ResNet \cite{He2016Deep} family. 
The performance gain in each case is noted in parentheses. 
$^\dagger$ indicates the numbers reported on the PyTorch page \cite{pytorch} and verified on our end. Whereas $^\ddagger$ indicates the numbers taken from \cite{Yun2019CutMix}. 
It should be noted that the numbers reported for baseline ResNets in \cite{Yun2019CutMix} are higher than PyTorch and have been noted below.
The numbers reported for all other approaches are taken from their respective papers. 
For consistent comparison we report the accuracy for all approaches where train and test resolution is fixed to $224\times224$ and specifically note the variants where test resolution deviates.
Averaging  the softmax (SM) scores of random crops (RCs) results in better performance than that of averaging at the deep feature vector (FV) level.
We also provide the numbers when 5 five fixed crops (FCs) suggested by AlexNet \cite{krizhevsky2012imagenet} are employed or additionally when mirrored versions of these fixed crops (\ie, MFCs) are also used.
Including the mirrored random crops (MRCs) in the averaging results in better performance. We conduct an ablation study (Tab \ref{table_resnet_supp}) demonstrating combining RCs, MRCs, FCs and MFCs results in the best achievable performance.  
The performance gain is high for smaller networks and low for larger ones. 
An increase in the number of crops results in enhanced performance, however, performance gain saturates after a certain number of crops as evidenced in Fig \ref{fig_resnet_performance_resnet18}.}
\resizebox{0.98\textwidth}{!}{
\label{table_resnet}
\centering
\setlength{\arrayrulewidth}{.1em}
\setlength\tabcolsep{3.6pt}
\small
\begin{tabular}{cc|c|c|c|c|c|c}
\multicolumn{2}{c|}{\multirow{5}{*}{\textbf{Approach}}} & \textbf{Custom Training} & \textbf{ResNet-18} & \textbf{ResNet-34} & \textbf{ResNet-50} & \textbf{ResNet-101} & \textbf{ResNet-152}\\ \cline{3-8}
& & Image Resized To & \multicolumn{5}{c}{256}\\
& & Input Crop Size & \multicolumn{5}{c}{$224\times224$}\\
& & 1.0 - (Crop to Image Ratio) & \multicolumn{5}{c}{0.125}\\ \cline{4-8}
& & Feature Dimension & \multicolumn{2}{c|}{512} & \multicolumn{3}{c}{2048}\\ \hline

& Center Crop$^\dagger$ & - & 69.76 & 73.31 &	76.13 & 77.37 & 78.31 \\ \hline
\multirow{4}{*}{\rotatebox[origin=b]{90}{\textbf{MID \scriptsize(RCs)}}} 
& \scriptsize(Average of FV of 10 RCs) & \textcolor{darkergreen}{\xmark} & $71.42^{\textbf{(+1.66)}}$ & $74.74^{\textbf{(+1.43)}}$ & $77.22^{\textbf{(+1.09)}}$ & $78.46^{\textbf{(+1.09)}}$ & $79.39^{\textbf{(+1.08)}}$ \\ 
 & \scriptsize(Average of FV of 20 RCs) & \textcolor{darkergreen}{\xmark} & $71.56^{\textbf{(+1.80)}}$ & $74.93^{\textbf{(+1.62)}}$ & $77.30^{\textbf{(+1.17)}}$ & $78.65^{\textbf{(+1.28)}}$ & $79.54^{\textbf{(+1.23)}}$ \\
 & \scriptsize(Average of SM of 10 RCs) & \textcolor{darkergreen}{\xmark} & $71.64^{\textbf{(+1.88)}}$ & $74.88^{\textbf{(+1.57)}}$ & $77.44^{\textbf{(+1.31)}}$ & $78.67^{\textbf{(+1.30)}}$ & $79.58^{\textbf{(+1.27)}}$ \\ 
 & \scriptsize(Average of SM of 20 RCs) & \textcolor{darkergreen}{\xmark} & $71.83^{\textbf{(+2.07)}}$ & $75.14^{\textbf{(+1.83)}}$ & $77.49^{\textbf{(+1.36)}}$ & $78.85^{\textbf{(+1.48)}}$ & $79.79^{\textbf{(+1.48)}}$ \\ \hline
 
 \multirow{8}{*}{\rotatebox[origin=b]{90}{\textbf{MID \scriptsize(RCs + MRCs)}}} 
 & \scriptsize(Average of FV of 5 RCs + 5 MRCs) & \textcolor{darkergreen}{\xmark} & $71.65^{\textbf{(+1.89)}}$ & $75.01^{\textbf{(+1.70)}}$ & $77.43^{\textbf{(+1.30)}}$ & $78.61^{\textbf{(+1.24)}}$ & $79.64^{\textbf{(+1.33)}}$ \\ 
 & \scriptsize(Average of FV of 10 RCs + 10 MRCs) & \textcolor{darkergreen}{\xmark} & $71.85^{\textbf{(+2.09)}}$ & $75.15^{\textbf{(+1.84)}}$ & $77.50^{\textbf{(+1.37)}}$ & $78.79^{\textbf{(+1.42)}}$ & $79.75^{\textbf{(+1.44)}}$ \\
 & \scriptsize(Average of FV of 15 RCs + 15 MRCs) & \textcolor{darkergreen}{\xmark} & $71.94^{\textbf{(+2.18)}}$ & $75.24^{\textbf{(+1.93)}}$ & $77.56^{\textbf{(+1.43)}}$ & $78.85^{\textbf{(+1.48)}}$ & $79.76^{\textbf{(+1.45)}}$ \\ 
 & \scriptsize(Average of FV of 20 RCs + 20 MRCs) & \textcolor{darkergreen}{\xmark} & $71.91^{\textbf{(+2.15)}}$ & $75.30^{\textbf{(+1.99)}}$ & $77.57^{\textbf{(+1.44)}}$ & $78.88^{\textbf{(+1.51)}}$ & $79.82^{\textbf{(+1.51)}}$ \\ 
 & \scriptsize(Average of SM of 5 RCs + 5 MRCs) & \textcolor{darkergreen}{\xmark} & $71.90^{\textbf{(+2.14)}}$ & $75.28^{\textbf{(+1.97)}}$ & $77.64^{\textbf{(+1.51)}}$ & $78.77^{\textbf{(+1.40)}}$ & $79.84^{\textbf{(+1.53)}}$ \\ 
 & \scriptsize(Average of SM of 10 RCs + 10 MRCs) & \textcolor{darkergreen}{\xmark} & $72.12^{\textbf{(+2.36)}}$ & $75.42^{\textbf{(+2.11)}}$ & $77.78^{\textbf{(+1.65)}}$ & $78.93^{\textbf{(+1.56)}}$ & $79.97^{\textbf{(+1.66)}}$ \\
 & \scriptsize(Average of SM of 15 RCs + 15 MRCs) & \textcolor{darkergreen}{\xmark} & $72.19^{\textbf{(+2.43)}}$ & $75.47^{\textbf{(+2.16)}}$ & $77.78^{\textbf{(+1.65)}}$ & $79.02^{\textbf{(+1.65)}}$ & $79.99^{\textbf{(+1.68)}}$ \\ 
 & \scriptsize(Average of SM of 20 RCs + 20 MRCs) & \textcolor{darkergreen}{\xmark} & $72.24^{\textbf{(+2.48)}}$ & $75.49^{\textbf{(+2.18)}}$ & $77.79^{\textbf{(+1.66)}}$ & $79.03^{\textbf{(+1.66)}}$ & $79.99^{\textbf{(+1.68)}}$ \\\hline
 
 \multirow{4}{*}{\rotatebox[origin=b]{90}{\scriptsize\textbf{FCs + MFCs}}} 
 & \scriptsize(Average of FV of 5 FCs) & \textcolor{darkergreen}{\xmark} & $71.31^{\textbf{(+1.55)}}$ & $74.78^{\textbf{(+1.47)}}$ & $77.12^{\textbf{(+0.99)}}$ & $78.66^{\textbf{(+1.29)}}$ & $79.40^{\textbf{(+1.09)}}$ \\ 
 & \scriptsize(Average of FV of 5 FCs + 5 MFCs) & \textcolor{darkergreen}{\xmark} & $71.85^{\textbf{(+2.09)}}$ & $75.27^{\textbf{(+1.96)}}$ & $77.44^{\textbf{(+1.31)}}$ & $78.93^{\textbf{(+1.56)}}$ & $79.73^{\textbf{(+1.42)}}$ \\ 
 & \scriptsize(Average of SM of 5 FCs) & \textcolor{darkergreen}{\xmark} & $71.70^{\textbf{(+1.94)}}$ & $75.09^{\textbf{(+1.78)}}$ & $77.35^{\textbf{(+1.22)}}$ & $78.84^{\textbf{(+1.47)}}$ & $79.69^{\textbf{(+1.38)}}$ \\ 
 & \scriptsize(Average of SM of 5 FCs + 5 MFCs) & \textcolor{darkergreen}{\xmark} & $72.23^{\textbf{(+2.47)}}$ & $75.63^{\textbf{(+2.32)}}$ & $77.66^{\textbf{(+1.53)}}$ & $79.15^{\textbf{(+1.78)}}$ & $80.01^{\textbf{(+1.70)}}$ \\\hline
 
& MEAL \cite{Shen2019MEAL}                        &       Training Required & - & - & 76.42 & - & - \\ 
& MEAL Plus \cite{Shen2019MEAL}                   &       Training Required & - & - & 78.21 & - & - \\
& MEAL V2 \cite{Shen2021MEAL}                     &       Training Required & 73.19 & - & 80.67 & - & - \\ \hline
& Baseline$^\ddagger$                             &       - & \multicolumn{2}{c|}{\multirow{5}{*}{N/A}} & 76.32 & 78.13 & \multirow{5}{*}{N/A} \\
& Cutout$^\ddagger$ \cite{DeVries2017Improved}    &       Training Required &\multicolumn{2}{c|}{} & 77.07 & 79.28 & \\
& Mixup$^\ddagger$ \cite{Zhang2018mixup}          &       Training Required & \multicolumn{2}{c|}{} & 77.42 & 79.48 & \\
& Manifold Mixup$^\ddagger$ \cite{Verma2019Manifold} &       Training Required & \multicolumn{2}{c|}{} & 77.50 & - & \\
& CutMix$^\ddagger$ \cite{Yun2019CutMix}          &       Training Required & \multicolumn{2}{c|}{} & 78.60 & 79.83 & \\\hline
\multirow{4}{*}{\rotatebox[origin=b]{90}{\textbf{FixRes}}} &\scriptsize{Base FixRes}&       Training Required & - & - & 77.0 & - & - \\ 
& \scriptsize$(\text{adaptation + augmentation})$   &       Training Required & - & - & 77.1 & - & - \\
& \scriptsize$(\text{adaptation + augmentation + @ 384})$   &       Training Required & - & - & 79.1 & - & - \\
& \scriptsize$(\text{adaptation + augmentation + @ 320 + Billion-scale})$   &       Training Required & - & - & 82.5 & - & - \\
\end{tabular}
}
\vspace{-0.05in}
\end{table*}

\begin{table*}[ht]
\centering
\caption{Comparison of \textbf{Top-1} validation accuracy on ImageNet dataset for MobileNet family \cite{Sandler2018MobileNetV2, Howard2019Searching}. The performance gain in each case is noted in parentheses. $\dagger$ indicates the numbers reported on the PyTorch page and verified on our end. Whereas $^\ddagger$ indicates the numbers taken from \cite{Shen2021MEAL}. The numbers reported for MobileNet baselines in \cite{Shen2021MEAL} do not match with PyTorch and explicitly copied here.} 
\resizebox{0.68\textwidth}{!}{
\label{table_mobile_net}
\begin{tabular}{c|c|c|c|c}
\hline
Approach & Custom Training & MobileNet V2 & MobileNet V3-Small & MobileNet V3-Large\\ \hline 
Feature Dimension & - & 1280 & 576 & 960\\
Image Resized To & - & 256 & 256 & 256\\
Input Crop Size & - & $224\times224$ & $224\times224$ & $224\times224$ \\
1.0 - (Crop to Image Ratio) & - & 0.125 & 0.125 & 0.125\\ \hline
Center Crop $\dagger$ & - & 71.88 & 67.67 & 74.04 \\ \hline
\textbf{MID} \scriptsize(Average of FV of 10 RCs) & \textcolor{darkergreen}{\xmark} & $73.55^{\textbf{(+1.67)}}$ & $69.79^{\textbf{(+2.12)}}$ & $75.38^{\textbf{(+1.34)}}$\\ 
\textbf{MID} \scriptsize(Average of FV of 20 RCs) & \textcolor{darkergreen}{\xmark} & $73.66^{\textbf{(+1.78)}}$ & $69.87^{\textbf{(+2.20)}}$ & $75.40^{\textbf{(+1.36)}}$\\
\textbf{MID} \scriptsize(Average of SM of 10 RCs) & \textcolor{darkergreen}{\xmark} & $73.76^{\textbf{(+1.88)}}$ & $70.06^{\textbf{(+2.39)}}$ & $75.52^{\textbf{(+1.48)}}$\\ 
\textbf{MID} \scriptsize(Average of SM of 20 RCs) & \textcolor{darkergreen}{\xmark} & $73.88^{\textbf{(+2.00)}}$ & $70.27^{\textbf{(+2.60)}}$ & $75.66^{\textbf{(+1.62)}}$ \\\hline
Baseline$^\ddagger$ & - & - & 67.40 & 75.20 \\
MEAL V2$^\ddagger$ \cite{Shen2021MEAL} & Training Required & - & 69.65 & 76.92 \\
\end{tabular}
}
\vspace{-0.05in}
\end{table*}

\begin{table*}[ht]
\centering
\caption{Comparison of \textbf{Top-1} validation accuracy on ImageNet dataset for EfficientNet \cite{Tan2019EfficientNet} family (tf\_efficientnet\_bx\_ap variant). The performance gain in each case is noted in parentheses. $\dagger$ indicates the numbers reported on the Timm's page \cite{rw2019timm1} and verified on our end. Numbers reported with $^\ddagger$ are taken from \cite{Touvron2020Fixing}. Since, FixEfficientNet re-trains networks for different resolutions; these test resolutions are noted in the last row.} 
\resizebox{0.98\textwidth}{!}{
\label{table_tf_efficient_net_ap}
\begin{tabular}{c|c|c|c|c|c|c|c|c|c|c}
\hline
Approach & Custom Training & B0 & B1 & B2 & B3 & B4 & B5 & B6 & B7 & B8\\ \hline 
Feature Dimension & - & 1280 & 1280 & 1408 & 1536 & 1792 & 2048 & 2304 & 2560 & 2816\\
Image Resized To & - & 256 & 272 & 292 & 332 & 412 & 488 & 562 & 632 & 704\\
Input Crop Size & - & $224\times224$ & $240\times240$ & $260\times260$ & $300\times300$ & $380\times380$ & $456\times456$ & $528\times528$ & $600\times600$ & $672\times672$ \\ 
1.0 - (Crop to Image Ratio) & - & 0.125 & 0.118 & 0.110 & 0.096 & 0.078 & 0.066 & 0.060 & 0.051 & 0.046\\\hline
Center Crop $\dagger$ & - & 77.09 & 79.28 &	80.30 & 81.82 & 83.25 & 84.25 &	84.79 &	85.12 &	85.37\\ \hline
\textbf{MID} \scriptsize(Average of FV of 10 RCs) & \textcolor{darkergreen}{\xmark} & $78.24^{\textbf{(+1.15)}}$ &	$80.10^{\textbf{(+0.82)}}$ &	$80.93^{\textbf{(+0.63)}}$ &	$82.32^{\textbf{(+0.50)}}$ &	$83.62^{\textbf{(+0.37)}}$ &	$84.52^{\textbf{(+0.27)}}$ &	$84.96^{\textbf{(+0.17)}}$ &	$85.25^{\textbf{(+0.13)}}$ &	$85.49^{\textbf{(+0.12)}}$\\ 
\textbf{MID} \scriptsize(Average of FV of 20 RCs) & \textcolor{darkergreen}{\xmark} & $78.32^{\textbf{(+1.23)}}$ &	$80.15^{\textbf{(+0.87)}}$ &	$80.97^{\textbf{(+0.67)}}$ &	$82.37^{\textbf{(+0.55)}}$ &	$83.68^{\textbf{(+0.43)}}$ &	$84.57^{\textbf{(+0.32)}}$ &	$84.96^{\textbf{(+0.17)}}$ &	$85.27^{\textbf{(+0.15)}}$ &	$85.50^{\textbf{(+0.13)}}$\\
\textbf{MID} \scriptsize(Average of SM of 10 RCs) & \textcolor{darkergreen}{\xmark} & $78.40^{\textbf{(+1.31)}}$ &	$80.24^{\textbf{(+0.96)}}$ &	$81.10^{\textbf{(+0.80)}}$ &	$82.41^{\textbf{(+0.59)}}$ &	$83.75^{\textbf{(+0.50)}}$ &	$84.62^{\textbf{(+0.37)}}$ &	$85.03^{\textbf{(+0.22)}}$ &	$85.32^{\textbf{(+0.20)}}$ &	$85.55^{\textbf{(+0.18)}}$\\ 
\textbf{MID} \scriptsize(Average of SM of 20 RCs) & \textcolor{darkergreen}{\xmark} & $78.43^{\textbf{(+1.34)}}$ &	$80.34^{\textbf{(+1.06)}}$ &	$81.16^{\textbf{(+0.86)}}$ &	$82.47^{\textbf{(+0.65)}}$ &	$83.79^{\textbf{(+0.54)}}$ &	$84.69^{\textbf{(+0.44)}}$ &	$85.06^{\textbf{(+0.25)}}$ &	$85.34^{\textbf{(+0.22)}}$ &	$85.58^{\textbf{(+0.21)}}$\\ \hline
EfficientNet AdvProp$^\ddagger$ \cite{Xie2020Adversarial}  &  - & 77.6 & 79.6 & 80.5 & 81.9 & 83.3 & 84.3 & 84.8 & 85.2 & 85.5\\ 
FixEfficientNet AdvProp$^\ddagger$ \cite{Touvron2020Fixing} & Training Required & 79.3 & 81.3 & 82.0 & 83.0 & 84.0 & 84.7 & 84.9 & 85.3 & 85.7\\ \hline
FixEfficientNet Test Res$^\ddagger$ \cite{Touvron2020Fixing} & - & 320 & 384 & 420 & 472 & 512 & 576 & 576 & 632 & 800 \\
\end{tabular}
}
\vspace{-0.05in}
\end{table*}

%-------------------------------------------------------------------------
\section{Experimental Settings}
\label{sec:Experiment}

\subsection{Data Set}
We conduct experiments on ILSVRC-2012 benchmark for classification \cite{Deng2009ImageNet} where models have been trained to discriminate among 1,000 classes using 1.2 million images in the training split. Since, we do not employ any training, we use 50,000 images in the validation split for  evaluation.    

\subsection{Networks}
We evaluated four different families of modern deep networks for object recognition. EfficientNet and NFNet are the best performing state-of-the-art models in supervised setting. Whereas ResNet and MobileNet architectures are of practical deployment importance due to their smaller memory footprint. For each model, we evaluated averaging at deep feature and softmax levels. Averaging at the logit level results in numbers identical to the ones resulting from averaging of feature vectors and not reported for brevity. For ResNet models (Tab \ref{table_resnet}), we have investigated fixed crops (FCs), mirrored fixed crops (MFCs), random crops (RCs), mirrored random crops (MRCs), and additionally combining RCs and MRCs (Tab \ref{table_resnet_supp}). For other models we constrained the evaluation to 10 or 20 random crops per image, and demonstrated performance gains on par with expensive training-based approaches.              

\paragraph{ResNet \cite{He2016Deep}}
We use the pre-trained models from PyTorch \cite{pytorch}, resize the input image to $256$ respecting the aspect ratio and then take $224\times224$ central or random crop. For resizing, bilinear interpolation is employed. Results available in Tab \ref{table_resnet}.       

\paragraph{MobileNet \cite{Sandler2018MobileNetV2, Howard2019Searching}}
For MobileNet V2 \cite{Sandler2018MobileNetV2} and V3 \cite{Howard2019Searching}, we use the pre-trained models from PyTorch \cite{pytorch}. For V3, both small and large models are evaluated. The input size, crops size, and the interpolation method stay the same as that for ResNet. Results are listed in Tab \ref{table_mobile_net}.     

\paragraph{EfficientNet \cite{Tan2019EfficientNet}} 
For EfficientNet architecture, we use the pre-trained models from Timm library \cite{rw2019timm} where weights have been converted from TensorFlow to PyTorch. Timm library provides the original pre-trained models \cite{Tan2019EfficientNet} as well as the NoisyStudent \cite{Xie2020Self-training} and AdvProp \cite{Xie2020Adversarial} variants. For our experiments we used four EfficientNet variants from Timm; identified by notation: efficientnet\_bx, tf\_efficientnet\_bx, tf\_efficientnet\_bx\_ns, and tf\_efficientnet\_bx\_ap where x can range from 0 to 8; tf, ns, and ap refers to TensorFlow, NoisyStudent, and AdvProp respectively. For efficientnet\_bx, only five pre-trained models (b0 through b4) are available. For each EfficientNet architecture, the input image is resized to the specific input size and then specific central or random crop is taken. Specific input and crop sizes are noted in respective tables. To comply with Timm \cite{rw2019timm}, we use bicubic interpolation. Results for tf\_efficientnet\_bx\_ap, tf\_efficientnet\_bx\_ns, efficientnet\_bx, and tf\_efficientnet\_bx are respectively available in Tabs \ref{table_tf_efficient_net_ap}, \ref{table_tf_efficient_net_ns}, \ref{table_efficient_net}, and \ref{table_tf_efficient_net}.     

\paragraph{NFNet \cite{Brock2021High}}
NFNet is the current state-of-the-art architecture for image recognition. Again we used pre-trained models from Timm \cite{rw2019timm}. Results for NFNet variants are available in Tab \ref{table_nfnet} with respective input and crop size information. Bicubic interpolation is used for resizing.     

%-------------------------------------------------------------------------
\section{Results}
\label{sec:Results}

\subsection{Comparison against SOTA}
\paragraph{ResNet}
Results for ResNet family are shown in Tab \ref{table_resnet}. We report Top-1 accuracy for each ResNet model as reported on PyTorch page \cite{pytorch} for single central-crop evaluation. For each ResNet model, we report the performance gain due to averaging of the feature vectors (FV) and softmax scores (SM), and use 10 or 20 random crops in each evaluation. We see a consistent improvement for each model with performance gain in Top-1 accuracy ranging from $\textbf{1.08\%}$ to $\textbf{2.07\%}$. The performance improves with increasing the number of random crops being averaged regardless of the underlying model. Additionally averaging the softmax scores results in better performance than averaging of feature vectors. We should further note that performance gain for smaller models (\eg, ResNet-18 and ResNet-34) is higher than the larger ones (\eg, ResNet-50, ResNet-101, and ResNet-152). We provide the comparison against several recent training-based approaches \cite{Shen2019MEAL, Shen2021MEAL, Touvron2019Fixing, Yun2019CutMix} and list whichever numbers are available for any of the five ResNet models. It should be noted that many such approaches chose to focus on ResNet-50. It is interesting to see that averaging of random crops can achieve comparable performance to some of the training-based methods. For examples, for ResNet-50, averaging is either outperforming (MEAL, Cutout, Mixup, FixRes, FixRes (adaptation + augmentation)) or comparable (Manifold Mixup) to training-based approaches which require re-training the model and/or exploiting datasets beyond conventional ImageNet-2012 train split.

Supplementing random crops with mirrored versions of additional random crops as well as using fixed crops and their flipped versions as originally suggested by AlexNet further boosts the performance. We conduct an ablation demonstrating that combining fixed and random crops results in the best achievable performance.

\begin{figure*}[ht!]
\centering
\includegraphics[width=0.80\linewidth]{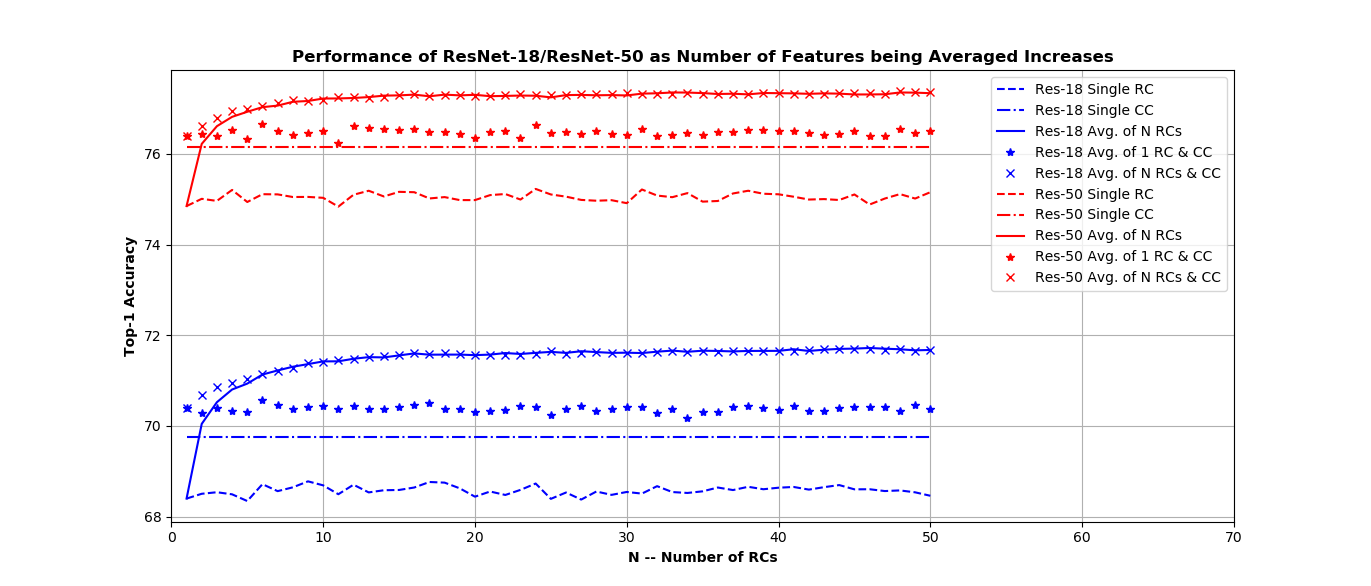}
\caption{Top-1 accuracy of ResNet-18/ResNet-50 as the function of number of RCs being averaged. When only a single crop is used per image, on average using central crop (CC) is better than RC. However as the number of RCs being averaged increases (even average of two RCs), the performance increases. The performance is saturated after about 20 RCs. We also demonstrate that including CC in averaging helps improving the performance when number of RCs per image is less than 5. In general using even a single RC in addition to the CC results in better performance than using just the central one.}
\label{fig_resnet_performance_resnet18}
\end{figure*}

\paragraph{MobileNet}
We document results for MobileNet V2 \cite{Sandler2018MobileNetV2}, and V3 \cite{Howard2019Searching} in Tab \ref{table_mobile_net} where we first list the numbers from PyTorch page \cite{pytorch} for V2 and two variants of V3. Averaging of random crops results in consistent improvement for all three MobileNet models with better performance gains for larger number of crops and softamx averaging outperforming feature averaging. The performance gain in Top-1 accuracy for MobileNet family ranges from $\textbf{1.34\%}$ to $\textbf{2.60\%}$. Similar to ResNet, performance gain for smaller model (V3-Small) is more than that for larger ones (V2 and V3-Larger). In Tab \ref{table_mobile_net}, we also report the numbers for one of the training-based approach \ie, MEAL V2 \cite{Shen2021MEAL}. The performance improvement for MEAL V2 over the baseline V3-Small and V3-Large is $\textbf{2.25\%}$ and $\textbf{1.72\%}$ respectively, whereas, averaging of random crops results in $\textbf{2.60\%}$ and $\textbf{1.62\%}$ gain in Top-1 accuracy \ie, outperforming MEAL V2 for V3-Small and comparable for V3-Large.                       

\paragraph{EfficientNet}
Results for EfficientNet AdvProp are available in Tab \ref{table_tf_efficient_net_ap} \& Fig \ref{fig_advprop}. The results for other EfficientNet variants can be found in Tabs \ref{table_efficient_net}, \ref{table_tf_efficient_net}, and \ref{table_tf_efficient_net_ns}. For AdvProp variant, the performance gain ranges from $\textbf{0.12\%}$ to $\textbf{1.34\%}$. We notice a larger gain for smaller models, and consistent with earlier results that larger number of crops and averaging at sofmax level helps. We provide a comparison against FixEfficientNet \cite{Touvron2020Fixing} which re-trains the underlying networks on test resolutions and employs additional data beyond ImageNet-2012 training set. It is interesting to note that for larger networks (\eg, B5 -- B8), averaging of random crops is comparable or outperforming FixEfficientNet, which raises the question if training such big models with additional million/billion scale data is worth the effort? Additionally as the crop to image ratio becomes larger for such bigger networks, the gains due to several random crop averaging diminishes as well, we explicitly discuss this phenomenon in the next subsection.                        

\paragraph{NFNet}
We report the results for NFNet family in Tab \ref{table_nfnet} where we see the Top-1 performance gain ranging from $\textbf{0.05\%}$ to $\textbf{0.63\%}$. The performance gain for NFNet is low compared to EfficientNet. This is partly due to larger crop to image ratio as there is relatively smaller area around the center crop from which a random crop can be chosen.

\subsection{Additional Results \& Analysis}
\paragraph{Number of Random Crops}
The number of random crops being averaged plays an important role in squeezing the performance out of the pre-trained models. Generally a larger number of random crops per image results in better performance, as seen in almost all results where irrespective of the underlying architecture, averaging of 20 crops resulted in better performance than averaging of 10. In Fig \ref{fig_resnet_performance_resnet18}, we study the performance gain as a function of number of crops being averaged for ResNet-18/ResNet-50 and a similar figure (Fig \ref{fig_resnet_performance}) for the complete ResNet family is also provided. 
It is apparent that including more crops in averaging results in better performance, however, performance saturates after about 20 random crops.      

\paragraph{Importance of Central Crop}
In general, central crop of a resized image provides a reasonable coverage of the object of interest. To study the importance of center crop, we included its representation in the average. As clear from Fig \ref{fig_resnet_performance_resnet18}, including central crop in averaging definitely helps when the number of random crops is less than 5. As the number of random crops increases, the importance of including center crop in averaging diminishes. It is also interesting to note that even including one extra random crop per image in addition to the conventional center crop results in better performance than just using the central crop.         

\paragraph{Averaging of Feature Vectors vs SoftMax Scores}
We have investigated averaging the representation of random crops at feature vector, logit, and softmax score levels. Averaging at the logit level results into identical performance as that of averaging at the deep feature level. Whereas averaging at the softmax level results in better performance than that of feature averaging as can be seen throughout the results (Tab. \ref{table_resnet} -- \ref{table_tf_efficient_net}). It is worth noting that averaging of softmax has also been investigated by others \cite{Shen2021MEAL} as a soft supervisory label to train better student networks.         

\paragraph{Smaller vs Larger Models} 
Overall the performance gain is higher for smaller models than for the larger ones. This is true regardless of the network family. This is partially due to the reason that larger input is used for larger networks (\eg,  B8 compared to B0). For ResNet family where the input crop size stays the same, we notice a larger gain for small models (\eg, ResNet-18) than larger ones (\eg, ResNet-152). Higher performance gain for smaller models and lower for larger ones is also consistent with training based methods \eg, \cite{Touvron2020Fixing} (Tabs \ref{table_tf_efficient_net_ap}, \ref{table_tf_efficient_net_ns}).

\begin{table*}[ht]
\centering
\caption{Comparison of \textbf{Top-1} validation accuracy on ImageNet dataset for NFNet \cite{Brock2021High} family. The performance gain in each case is noted in parentheses. $\dagger$ indicates the numbers reported on the Timm's page \cite{rw2019timm1} and verified on our end. Softmax (SM) scores averaging is better than feature (FV) averaging. The performance gain is high for smaller networks and low for larger ones.}  
\resizebox{0.88\textwidth}{!}{
\label{table_nfnet}
\begin{tabular}{c|c|c|c|c|c|c|c|c}
\hline
Approach & Custom Training & F0 & F1 & F2 & F3 & F4 & F5 & F6 \\ \hline
Feature Dimension & - & 3072 & 3072 & 3072 & 3072 & 3072 & 3072 & 3072\\
Image Resized To & - & 284 & 351 & 382 & 443 & 538 & 570 & 602\\
Input Crop Size & - & $256\times256$ & $320\times320$ & $352\times352$ & $416\times416$ & $512\times512$ & $544\times544$ & $576\times576$\\ 
1.0 - (Crop to Image Ratio) & - & 0.099 & 0.088 & 0.079 & 0.061 & 0.048 & 0.046 & 0.043\\\hline

Center Crop $\dagger$ & - & 83.34 &	84.60 &	84.99 &	85.56 & 85.66 & 85.71 &	86.30\\ \hline
\textbf{MID} \scriptsize(Average of FV of 10 RCs) & \textcolor{darkergreen}{\xmark} & $83.67^{\textbf{(+0.33)}}$ & $84.79^{\textbf{(+0.19)}}$ & $85.13^{\textbf{(+0.14)}}$ & $85.79^{\textbf{(+0.23)}}$ & $85.84^{\textbf{(+0.18)}}$ & $85.91^{\textbf{(+0.20)}}$ & $86.35^{\textbf{(+0.05)}}$\\ 
\textbf{MID} \scriptsize(Average of FV of 20 RCs) & \textcolor{darkergreen}{\xmark} & $83.73^{\textbf{(+0.39)}}$ & $84.81^{\textbf{(+0.21)}}$ & $85.13^{\textbf{(+0.14)}}$ & $85.82^{\textbf{(+0.26)}}$ & $85.85^{\textbf{(+0.19)}}$ & $85.94^{\textbf{(+0.23)}}$ & $86.33^{\textbf{(+0.03)}}$\\
\textbf{MID} \scriptsize(Average of SM of 10 RCs) & \textcolor{darkergreen}{\xmark} & $83.77^{\textbf{(+0.53)}}$ & $84.90^{\textbf{(+0.30)}}$ & $85.22^{\textbf{(+0.23)}}$ & $85.87^{\textbf{(+0.31)}}$ & $85.91^{\textbf{(+0.25)}}$ & $85.95^{\textbf{(+0.24)}}$ & $86.43^{\textbf{(+0.13)}}$\\ 
\textbf{MID} \scriptsize(Average of SM of 20 RCs) & \textcolor{darkergreen}{\xmark} & $83.87^{\textbf{(+0.63)}}$ & $84.92^{\textbf{(+0.32)}}$ & $85.20^{\textbf{(+0.21)}}$ & $85.89^{\textbf{(+0.33)}}$ & $85.90^{\textbf{(+0.24)}}$ & $86.02^{\textbf{(+0.31)}}$ & $86.41^{\textbf{(+0.11)}}$\\ \hline
\end{tabular}
}
\vspace{-0.05in}
\end{table*}

\paragraph{Importance of Crop to Image Ratio}
Another reason behind better performance for smaller models is due to crop to image size ratio which we note for each model and all the network families in their respective tables. When this ratio is smaller, there is more space for the selection of random crop, and selecting more random crops provides better coverage of the underlying object. This can be noticed for EfficientNet (Tabs \ref{table_tf_efficient_net_ap}, \ref{table_tf_efficient_net_ns}, \ref{table_efficient_net}, \ref{table_tf_efficient_net}) and NFNet (Tab \ref{table_nfnet}) where smaller models like B0 and F0 have smaller crop to image ratio than larger ones like B7/B8 and F6. Comparing the performance gains for NFNet (Tab \ref{table_nfnet}) against EfficientNet (Tab \ref{table_tf_efficient_net_ap}), we can see comparatively larger gains for EfficientNet due to relatively smaller crop to image ratios \eg, 0.875 for B0 compared to 0.901 for F0. Similarly, comparing efficientnet\_bx variant (Tab \ref{table_efficient_net}) against other variants specifically tf\_efficientnet\_bx\_ns (Tab \ref{table_tf_efficient_net_ns}), and tf\_efficientnet\_bx\_ap (Tab \ref{table_tf_efficient_net_ap}), we notice smaller gains especially for B1 -- B4 as the images are resized to the actual crop size and wiggle room to choose random crops is only available in one dimension. Another related issue is the size of the underlying image after resizing as can be seen in Fig \ref{fig_longer_side_dist} where number of images in ILSVRC-2012 val split having longer dimension larger than 500 pixels are relatively fewer. That is why models that resize an image to more than 500 pixels (\eg, B6 -- B8, F4 -- F6, and L2) do not gain much due to averaging of random crops as there is not much room.                  

\begin{figure}[ht!]
\centering
\includegraphics[width=0.85\linewidth]{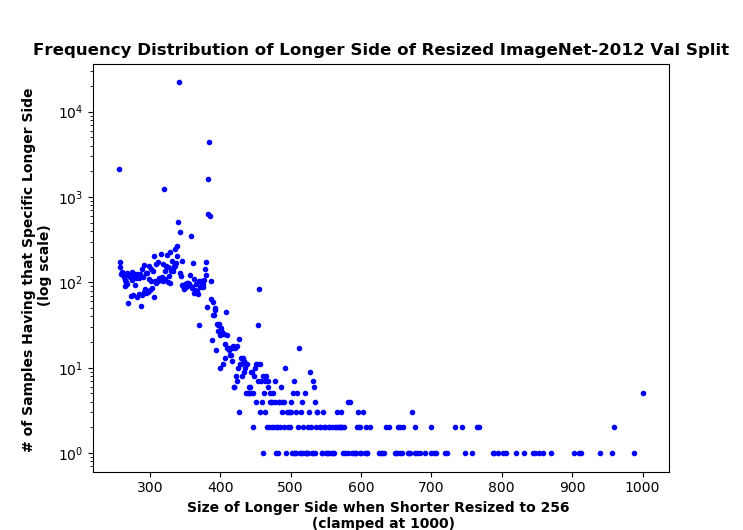}
\caption{Frequency distribution of longer side of images in ILSVRC-2012 val split when shorter side is resized to 256 pixels. Noticeably many images are larger than 256 pixels, and have a good chance of benefiting from averaging of RCs. We should further note that after resizing, fewer images have longer sides greater than 400 pixels, so a strategy to choose number of crops as a function of image size is effective both computationally and performance-wise.}
\label{fig_longer_side_dist}
\end{figure}

\paragraph{Results for Other EfficientNet Variants}
The results for other EfficientNet variants \ie, efficientnet\_bx, tf\_efficientnet\_bx, and tf\_efficientnet\_bx\_ns are available in Tabs \ref{table_efficient_net}, \ref{table_tf_efficient_net}, and \ref{table_tf_efficient_net_ns} respectively. For best performing EfficientNet variants \ie, NoisyStudent \cite{Xie2020Self-training} and AdvProp \cite{Xie2020Adversarial}, we are able to achieve comparable performance to that of FixEfficientNet \cite{Touvron2020Fixing} without requiring expensive million scale training. The visual performance comparison for AdvProp and NoisyStudent variants against FixEfficientNet versions is available in Figs \ref{fig_advprop} and \ref{fig_noisystudent} respectively. We should note for efficientnet\_bx only four pre-trained models (B0 -- B4) are available from Timm \cite{rw2019timm}. For tf\_efficientnet\_bx variant, only B0 comparison (\ie, MEAL V2 \cite{Shen2021MEAL}) is available and has been listed in the respective table (Tab \ref{table_tf_efficient_net}).

\begin{figure}[ht!]
\centering
\includegraphics[width=.98\linewidth]{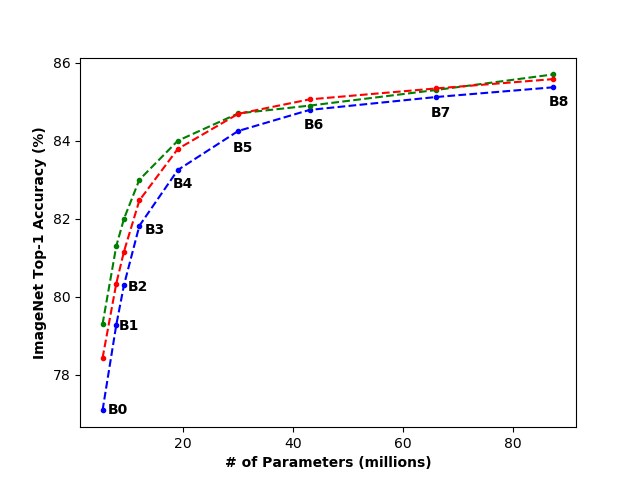}
\caption{Performance comparison of EfficienNet AdvProp \cite{Xie2020Adversarial} compared to baselines (\textcolor{bluecol}{blue curve}) reported by Timm \cite{rw2019timm1}. Our performance (\textcolor{redcol}{red curve}) is comparable to training-based FixEfficientNet \cite{Touvron2020Fixing} (\textcolor{greencol}{green curve}).}  
\label{fig_advprop}
\end{figure}

\begin{figure*}[ht!]
\centering
\includegraphics[width=.78\linewidth]{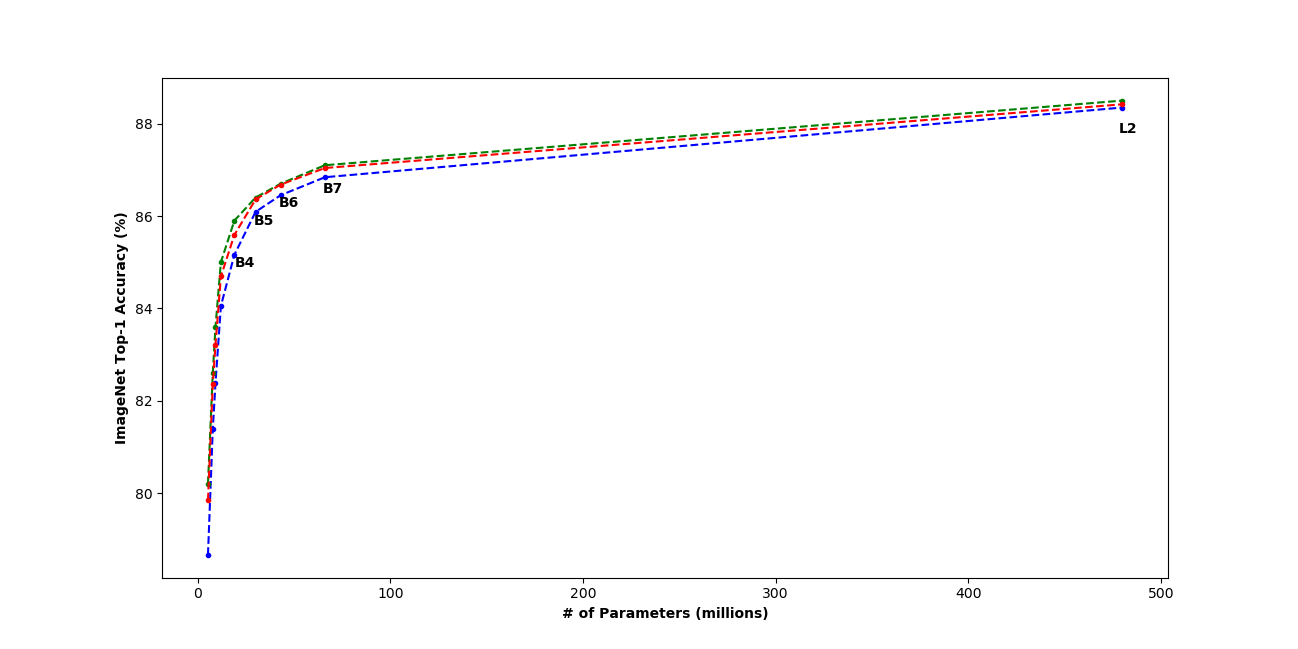}
\caption{Performance comparison of EfficienNet NoisyStudent \cite{Xie2020Self-training} compared to baselines (\textcolor{bluecol}{blue curve}) reported by Timm \cite{rw2019timm1}. Our performance (\textcolor{redcol}{red curve}) is comparable to training-based FixEfficientNet \cite{Touvron2020Fixing} (\textcolor{greencol}{green curve}).}  
\label{fig_noisystudent}
\end{figure*}

\begin{table*}[ht!]
\centering
\caption{Comparison of \textbf{Top-1} validation accuracy on ImageNet dataset for EfficientNet \cite{Tan2019EfficientNet} family (tf\_efficientnet\_bx\_ns variant). The performance gain in each case is noted in parentheses. $\dagger$ indicates the numbers reported on the Timm's page \cite{rw2019timm1} and verified on our end. Numbers reported with $^\ddagger$ are taken from \cite{Touvron2020Fixing}. Since, FixEfficientNet re-trains networks for different resolutions; these test resolutions are noted in the last row. Performing averaging of the softmax (SM) scores of random crops (RCs) results in better performance than that of averaging at the deep feature vector (FV) level. The performance gain is high for smaller networks and low for larger ones.} 
\resizebox{0.98\textwidth}{!}{
\label{table_tf_efficient_net_ns}
\begin{tabular}{c|c|c|c|c|c|c|c|c|c|c}
\hline
Approach & Custom Training & B0 & B1 & B2 & B3 & B4 & B5 & B6 & B7 & L2\\ \hline 
Feature Dimension & - & 1280 & 1280 & 1408 & 1536 & 1792 & 2048 & 2304 & 2560 & 5504\\
Image Resized To & - & 256 & 272 & 292 & 332 & 412 & 488 & 562 & 632 & 833\\
Input Crop Size & - & $224\times224$ & $240\times240$ & $260\times260$ & $300\times300$ & $380\times380$ & $456\times456$ & $528\times528$ & $600\times600$ & $800\times800$\\ 
1.0 - (Crop to Image Ratio) & - & 0.125 & 0.118 & 0.110 & 0.096 & 0.078 & 0.066 & 0.060 & 0.051 & 0.038\\\hline
Center Crop $\dagger$ & - & 78.66 & 81.39 &	82.38 & 84.05 & 85.16 & 86.09 &	86.45 &	86.84 &	88.35\\ \hline
\textbf{Ours} \scriptsize(Average of FV of 10 RCs) & \textcolor{darkergreen}{\xmark} & $79.55^{\textbf{(+0.89)}}$ &	$82.05^{\textbf{(+0.66)}}$ &	$82.92^{\textbf{(+0.54)}}$ &	$84.52^{\textbf{(+0.47)}}$ &	$85.41^{\textbf{(+0.25)}}$ &	$86.27^{\textbf{(+0.18)}}$ &	$86.62^{\textbf{(+0.17)}}$ &	$86.92^{\textbf{(+0.08)}}$ &	$88.35^{\textbf{(+0.00)}}$\\ 
\textbf{Ours} \scriptsize(Average of FV of 20 RCs) & \textcolor{darkergreen}{\xmark} & $79.65^{\textbf{(+0.99)}}$ &	$82.16^{\textbf{(+0.77)}}$ &	$83.05^{\textbf{(+0.67)}}$ &	$84.58^{\textbf{(+0.53)}}$ &	$85.48^{\textbf{(+0.32)}}$ &	$86.26^{\textbf{(+0.17)}}$ &	$86.60^{\textbf{(+0.15)}}$ &	$86.96^{\textbf{(+0.12)}}$ &	$88.40^{\textbf{(+0.05)}}$\\
\textbf{Ours} \scriptsize(Average of SM of 10 RCs) & \textcolor{darkergreen}{\xmark} & $79.76^{\textbf{(+1.10)}}$ &	$82.26^{\textbf{(+0.87)}}$ &	$83.10^{\textbf{(+0.72)}}$ &	$84.68^{\textbf{(+0.63)}}$ &	$85.53^{\textbf{(+0.37)}}$ &	$86.33^{\textbf{(+0.24)}}$ &	$86.67^{\textbf{(+0.20)}}$ &	$87.0^{\textbf{(+0.16)}}$ &	$88.39^{\textbf{(+0.04)}}$\\ 
\textbf{Ours} \scriptsize(Average of SM of 20 RCs) & \textcolor{darkergreen}{\xmark} & $79.84^{\textbf{(+1.18)}}$ &	$82.36^{\textbf{(+0.97)}}$ &	$83.21^{\textbf{(+0.83)}}$ &	$84.69^{\textbf{(+0.64)}}$ &	$85.59^{\textbf{(+0.43)}}$ &	$86.36^{\textbf{(+0.27)}}$ &	$86.68^{\textbf{(+0.21)}}$ &	$87.04^{\textbf{(+0.20)}}$ &	$88.42^{\textbf{(+0.07)}}$\\ \hline
EfficientNet NoisyStudent$^\ddagger$ \cite{Xie2020Self-training}  &  -  & 78.8 & 81.5 & 82.4 & 84.1 & 85.3 & 86.1 & 86.4 & 86.9 & 88.4\\ 
FixEfficientNet NoisyStudent$^\ddagger$ \cite{Touvron2020Fixing} & Training Required & 80.2 & 82.6 & 83.6 & 85.0 & 85.9 & 86.4 & 86.7 & 87.1 & 88.5\\ \hline
FixEfficientNet Test Res$^\ddagger$ \cite{Touvron2020Fixing} & - & 320 & 384 & 420 & 472 & 472 & 576 & 680 & 632 & 600\\
\end{tabular}
}
\vspace{-0.05in}
\end{table*}

\begin{table*}[ht!]
\centering
\caption{Comparison of \textbf{Top-1} validation accuracy on ImageNet dataset for EfficientNet \cite{Tan2019EfficientNet} family (efficientnet\_bx variant). It should be noted only five pre-trained models (B0 -- B4) are available for this variant. The performance gain in each case is noted in parentheses. $\dagger$ indicates the numbers reported on the Timm's page \cite{rw2019timm1} and verified on our end. Performing averaging of the softmax (SM) scores of random crops (RCs) results in better performance than that of averaging at the deep feature vector (FV) level. The performance gain is high for smaller networks and low for larger ones.} 
\resizebox{0.78\textwidth}{!}{
\label{table_efficient_net}
\begin{tabular}{c|c|c|c|c|c|c}
\hline
Approach & Custom Training & B0 & B1 & B2 & B3 & B4 \\ \hline 
Feature Dimension & - & 1280 & 1280 & 1408 & 1536 & 1792 \\
Image Resized To & - & 256 & 256 & 288 & 320 & 384\\
Input Crop Size & - & $224\times224$ & $256\times256$ & $288\times288$ & $320\times320$ & $384\times384$\\ 
1.0 - (Crop to Image Ratio) & - & 0.125 & 0.000 & 0.000 & 0.000 & 0.000\\ \hline
Center Crop $\dagger$ & - & 77.70 & 78.80 & 80.61 & 82.24 & 83.43 \\ \hline
\textbf{Ours} \scriptsize(Average of FV of 10 RCs) & \textcolor{darkergreen}{\xmark} & $78.59^{\textbf{(+0.89)}}$ & $79.37^{\textbf{(+0.57)}}$ & $81.01^{\textbf{(+0.40)}}$ & $82.49^{\textbf{(+0.25)}}$ & $83.54^{\textbf{(+0.11)}}$\\ 
\textbf{Ours} \scriptsize(Average of FV of 20 RCs) & \textcolor{darkergreen}{\xmark} & $78.68^{\textbf{(+0.98)}}$ & $79.34^{\textbf{(+0.54)}}$ & $81.05^{\textbf{(+0.44)}}$ & $82.49^{\textbf{(+0.25)}}$ & $83.63^{\textbf{(+0.20)}}$\\
\textbf{Ours} \scriptsize(Average of SM of 10 RCs) & \textcolor{darkergreen}{\xmark} & $78.88^{\textbf{(+1.18)}}$ & $79.46^{\textbf{(+0.66)}}$ &	$81.12^{\textbf{(+0.51)}}$ &	$82.60^{\textbf{(+0.36)}}$ &	$83.62^{\textbf{(+0.19)}}$\\ 
\textbf{Ours} \scriptsize(Average of SM of 20 RCs) & \textcolor{darkergreen}{\xmark} & $78.99^{\textbf{(+1.29)}}$ & $79.46^{\textbf{(+0.66)}}$ & $81.19^{\textbf{(+0.58)}}$ &	$82.58^{\textbf{(+0.34)}}$ & $83.70^{\textbf{(+0.27)}}$\\ \hline
\end{tabular}
}
\vspace{-0.05in}
\end{table*}

\begin{table*}[ht!]
\centering
\caption{Comparison of \textbf{Top-1} validation accuracy on ImageNet dataset for EfficientNet \cite{Tan2019EfficientNet} family (tf\_efficientnet\_bx variant). The performance gain in each case is noted in parentheses. $\dagger$ indicates the numbers reported on the Timm's page \cite{rw2019timm1} and verified on our end. Numbers reported for other approaches are taken from their respective papers. Performing averaging of the softmax (SM) scores of random crops (RCs) results in better performance than that of averaging at the deep feature vector (FV) level. The performance gain is high for smaller networks and low for larger ones.} 
\resizebox{0.98\textwidth}{!}{
\label{table_tf_efficient_net}
\begin{tabular}{c|c|c|c|c|c|c|c|c|c|c}
\hline
Approach & Custom Training & B0 & B1 & B2 & B3 & B4 & B5 & B6 & B7 & B8\\ \hline 
Feature Dimension & - & 1280 & 1280 & 1408 & 1536 & 1792 & 2048 & 2304 & 2560 & 2816\\
Image Resized To & - & 256 & 272 & 292 & 332 & 412 & 488 & 562 & 632 & 704\\
Input Crop Size & - & $224\times224$ & $240\times240$ & $260\times260$ & $300\times300$ & $380\times380$ & $456\times456$ & $528\times528$ & $600\times600$ & $672\times672$ \\
1.0 - (Crop to Image Ratio) & - & 0.125 & 0.118 & 0.110 & 0.096 & 0.078 & 0.066 & 0.060 & 0.051 & 0.046\\ \hline
Center Crop $\dagger$ & - & 76.85 &	78.83 &	80.09 &	81.64 &	83.02 &	83.81 &	84.11 &	84.94 &	85.37 \\ \hline
\textbf{Ours} \scriptsize(Average of FV of 10 RCs) & \textcolor{darkergreen}{\xmark} & $77.78^{\textbf{(+0.93)}}$ & $79.56^{\textbf{(+0.73)}}$ & $80.75^{\textbf{(+0.66)}}$ & $82.14^{\textbf{(+0.50)}}$ & $83.29^{\textbf{(+0.27)}}$ & $84.03^{\textbf{(+0.22)}}$ & $84.27^{\textbf{(+0.16)}}$ & $85.06^{\textbf{(+0.12)}}$ &	$85.52^{\textbf{(+0.15)}}$\\ 
\textbf{Ours} \scriptsize(Average of FV of 20 RCs) & \textcolor{darkergreen}{\xmark} & $77.83^{\textbf{(+0.98)}}$ & $79.72^{\textbf{(+0.89)}}$ & $80.87^{\textbf{(+0.78)}}$ & $82.16^{\textbf{(+0.52)}}$ & $83.34^{\textbf{(+0.32)}}$ & $84.11^{\textbf{(+0.30)}}$ & $84.33^{\textbf{(+0.22)}}$ & $85.10^{\textbf{(+0.16)}}$ &	$85.53^{\textbf{(+0.16)}}$\\
\textbf{Ours} \scriptsize(Average of SM of 10 RCs) & \textcolor{darkergreen}{\xmark} & $77.98^{\textbf{(+1.13)}}$ & $79.73^{\textbf{(+0.90)}}$ & $80.91^{\textbf{(+0.82)}}$ & $82.23^{\textbf{(+0.59)}}$ & $83.42^{\textbf{(+0.40)}}$ & $84.14^{\textbf{(+0.33)}}$ & $84.38^{\textbf{(+0.27)}}$ & $85.14^{\textbf{(+0.20)}}$ &	$85.60^{\textbf{(+0.23)}}$\\ 
\textbf{Ours} \scriptsize(Average of SM of 20 RCs) & \textcolor{darkergreen}{\xmark} & $78.03^{\textbf{(+1.18)}}$ & $79.89^{\textbf{(+1.06)}}$ & $81.05^{\textbf{(+0.96)}}$ & $82.29^{\textbf{(+0.65)}}$ & $83.45^{\textbf{(+0.43)}}$ & $84.20^{\textbf{(+0.39)}}$ & $84.45^{\textbf{(+0.34)}}$ & $85.21^{\textbf{(+0.27)}}$ &	$85.58^{\textbf{(+0.21)}}$\\ \hline
MEAL V2 \cite{Shen2021MEAL}                     &       Training Required & 78.29 & - & - & - & - & - & - & - &	-\\ \hline

\end{tabular}
}
\vspace{-0.05in}
\end{table*}

\paragraph{Ablation: Mirrored \& Fixed Crops}
Using ResNet models as the base, in Tab \ref{table_resnet_supp} we have investigated various additional settings. As reported in Tab \ref{table_resnet}; in addition to using just the random crops (RCs) only, we have explored using mirrored random crops (MRCs). While keeping the number of total crops fixed to 10 or 20, and comparing (i) using only RCs versus (ii) using a combination of RCs and MRCs, the lateral always results in better performance regardless of averaging at softmax or feature level. Focusing on ResNet-18 feature averaging and having the number of crops fixed to 10 and 20, a mixture of RCs and MRCs results in better performance (71.65 \& 71.85) compared to using RCs only (71.42 \& 71.56). This directly confirms our hypothesis of matching the inference-time distribution with that of training; as during training not only the RCs are employed but generally MRCs are also incorporated with a probability of 0.5.

AlexNet \cite{krizhevsky2012imagenet} originally suggested to use 5 fixed crops (FCs) -- 1 central and 4 corners as an attempt to match the train-test distribution. Although FCs might be sufficient for majority of square images; they may not work for elongated ones. We combined 5 FCs with an increased number of RCs (5, 10, 15, and 20) which resulted into consistent better performance (Tab \ref{table_resnet_supp}). Continuing with ResNet-18 feature averaging examples, we can see that including 5 \& 10 RCs in the average resulted into Top-1 accuracy of 71.49 and 71.59 compared to using only 5 FCs (71.31). This trend holds for larger ResNet models and averaging at the softmax layer.     

To further bridge the distributional gap, we also investigated using the mirrored versions of the fixed crops (MFCs) and additional mirrored random crops. We should note that AlexNet suggested to use the flipped versions of the same fixed crops whereas the flipped versions in our approach are not necessarily of the same random crops and hence even better align with the train-time distribution. We have identified the best achieved Top-1 performance for all ResNet models in bold font in \ref{table_resnet_supp} where we can see an average of fixed, random, mirrored-fixed, and mirrored-random crops have resulted in the best numbers.

\begin{table*}[ht!]
\centering
\caption{
Comparison of \textbf{Top-1} validation accuracy on ImageNet dataset for ResNet \cite{He2016Deep} family. 
$^\dagger$ indicates the numbers reported on the PyTorch page \cite{pytorch} and verified on our end. Compared to fixed crops (FCs) and mirrored fixed crops (MFCs), adding RCs and MRCs into the average results in enhanced performance. The best achieved performance (highlighted in bold) results when a mix of FCs, MFCs, RCs and MRCs is used. This is due to the fact that fixed crops may be sufficient for square images but the elongated images can further benefit from RCs and MRCs that provide additional coverage of the underlying object. Additionally since the MRCs are not necessarily the flipped versions of the RCs, rather independent, so they further bridge the train-test distributional gap.      
}
\resizebox{0.98\textwidth}{!}{
\label{table_resnet_supp}
\centering
\setlength{\arrayrulewidth}{.1em}
\setlength\tabcolsep{3.6pt}
\small
\begin{tabular}{cc|c|c|c|c|c|c}
\multicolumn{2}{c|}{\multirow{5}{*}{\textbf{Approach}}} & \textbf{Custom Training} & \textbf{ResNet-18} & \textbf{ResNet-34} & \textbf{ResNet-50} & \textbf{ResNet-101} & \textbf{ResNet-152}\\ \cline{3-8}
& & Image Resized To & \multicolumn{5}{c}{256}\\
& & Input Crop Size & \multicolumn{5}{c}{$224\times224$}\\
& & 1.0 - (Crop to Image Ratio) & \multicolumn{5}{c}{0.125}\\ \cline{4-8}
& & Feature Dimension & \multicolumn{2}{c|}{512} & \multicolumn{3}{c}{2048}\\ \hline

& Center Crop$^\dagger$ & - & 69.76 & 73.31 &	76.13 & 77.37 & 78.31 \\ \hline

 \multirow{4}{*}{\rotatebox[origin=b]{90}{\textbf{AlexNet}}} 
 & \scriptsize(Average of FV of 5 FCs) & \textcolor{darkergreen}{\xmark} & $71.31$ & $74.78$ & $77.12$ & $78.66$ & $79.40$ \\ 
 & \scriptsize(Average of FV of 5 FCs + 5 MFCs) & \textcolor{darkergreen}{\xmark} & $71.85$ & $75.27$ & $77.44$ & $78.93$ & $79.73$ \\\cline{3-8} 
 & \scriptsize(Average of SM of 5 FCs) & \textcolor{darkergreen}{\xmark} & $71.70$ & $75.09$ & $77.35$ & $78.84$ & $79.69$ \\ 
 & \scriptsize(Average of SM of 5 FCs + 5 MFCs) & \textcolor{darkergreen}{\xmark} & $72.23$ & $75.63$ & $77.66$ & $79.15$ & $80.01$ \\\hline
 
 \multirow{16}{*}{\rotatebox[origin=b]{90}{\textbf{AlexNet + Ours}}} 
 & \scriptsize(Average of FV of 5 FCs + 5 RCs) & \textcolor{darkergreen}{\xmark} & $71.49$ & $74.99$ & $77.25$ & $78.69$ & $79.56$ \\
 & \scriptsize(Average of FV of 5 FCs + 10 RCs) & \textcolor{darkergreen}{\xmark} & $71.59$ & $75.04$ & $77.33$ & $78.71$ & $79.63$ \\
 & \scriptsize(Average of FV of 5 FCs + 15 RCs) & \textcolor{darkergreen}{\xmark} & $71.63$ & $75.00$ & $77.41$ & $78.78$ & $79.63$ \\
 & \scriptsize(Average of FV of 5 FCs + 20 RCs) & \textcolor{darkergreen}{\xmark} & $71.64$ & $75.02$ & $77.40$ & $78.80$ & $79.62$ \\
 & \scriptsize(Average of FV of 5 FCs + 5 MFCs + 5 RCs + 5 MRCs) & \textcolor{darkergreen}{\xmark} & $71.97$ & $75.32$ & $77.67$ & $78.98$ & $79.84$ \\
 & \scriptsize(Average of FV of 5 FCs + 5 MFCs + 10 RCs + 10 MRCs) & \textcolor{darkergreen}{\xmark} & $71.99$ & $75.34$ & $77.67$ & $78.97$ & $79.89$ \\
 & \scriptsize(Average of FV of 5 FCs + 5 MFCs + 15 RCs + 15 MRCs) & \textcolor{darkergreen}{\xmark} & $71.97$ & $75.33$ & $77.60$ & $79.02$ & $79.87$ \\
 & \scriptsize(Average of FV of 5 FCs + 5 MFCs + 20 RCs + 20 MRCs) & \textcolor{darkergreen}{\xmark} & $71.96$ & $75.37$ & $77.66$ & $78.98$ & $79.83$ \\\cline{3-8}
 
 & \scriptsize(Average of SM of 5 FCs + 5 RCs) & \textcolor{darkergreen}{\xmark} & $71.87$ & $75.28$ & $77.45$ & $78.92$ & $79.81$ \\
 & \scriptsize(Average of SM of 5 FCs + 10 RCs) & \textcolor{darkergreen}{\xmark} & $71.96$ & $75.31$ & $77.55$ & $78.96$ & $79.85$ \\
 & \scriptsize(Average of SM of 5 FCs + 15 RCs) & \textcolor{darkergreen}{\xmark} & $72.01$ & $75.27$ & $77.64$ & $79.02$ & $79.86$ \\
 & \scriptsize(Average of SM of 5 FCs + 20 RCs) & \textcolor{darkergreen}{\xmark} & $71.98$ & $75.28$ & $77.63$ & $79.04$ & $79.89$ \\
 & \scriptsize(Average of SM of 5 FCs + 5 MFCs + 5 RCs + 5 MRCs) & \textcolor{darkergreen}{\xmark} & $72.36$ & $\textbf{75.66}$ & $77.83$ & $79.13$ & $80.14$ \\
 & \scriptsize(Average of SM of 5 FCs + 5 MFCs + 10 RCs + 10 MRCs) & \textcolor{darkergreen}{\xmark} & $72.36$ & $75.60$ & $77.87$ & $\textbf{79.18}$ & $\textbf{80.14}$ \\
 & \scriptsize(Average of SM of 5 FCs + 5 MFCs + 15 RCs + 15 MRCs) & \textcolor{darkergreen}{\xmark} & $\textbf{72.40}$ & $75.60$ & $77.86$ & $79.16$ & $80.13$ \\
 & \scriptsize(Average of SM of 5 FCs + 5 MFCs + 20 RCs + 20 MRCs) & \textcolor{darkergreen}{\xmark} & $72.33$ & $75.61$ & $\textbf{77.88}$ & $79.17$ & $80.11$ \\\hline

\end{tabular}
}
\vspace{-0.05in}
\end{table*}

\begin{figure*}[ht!]
\centering
\includegraphics[width=.88\linewidth]{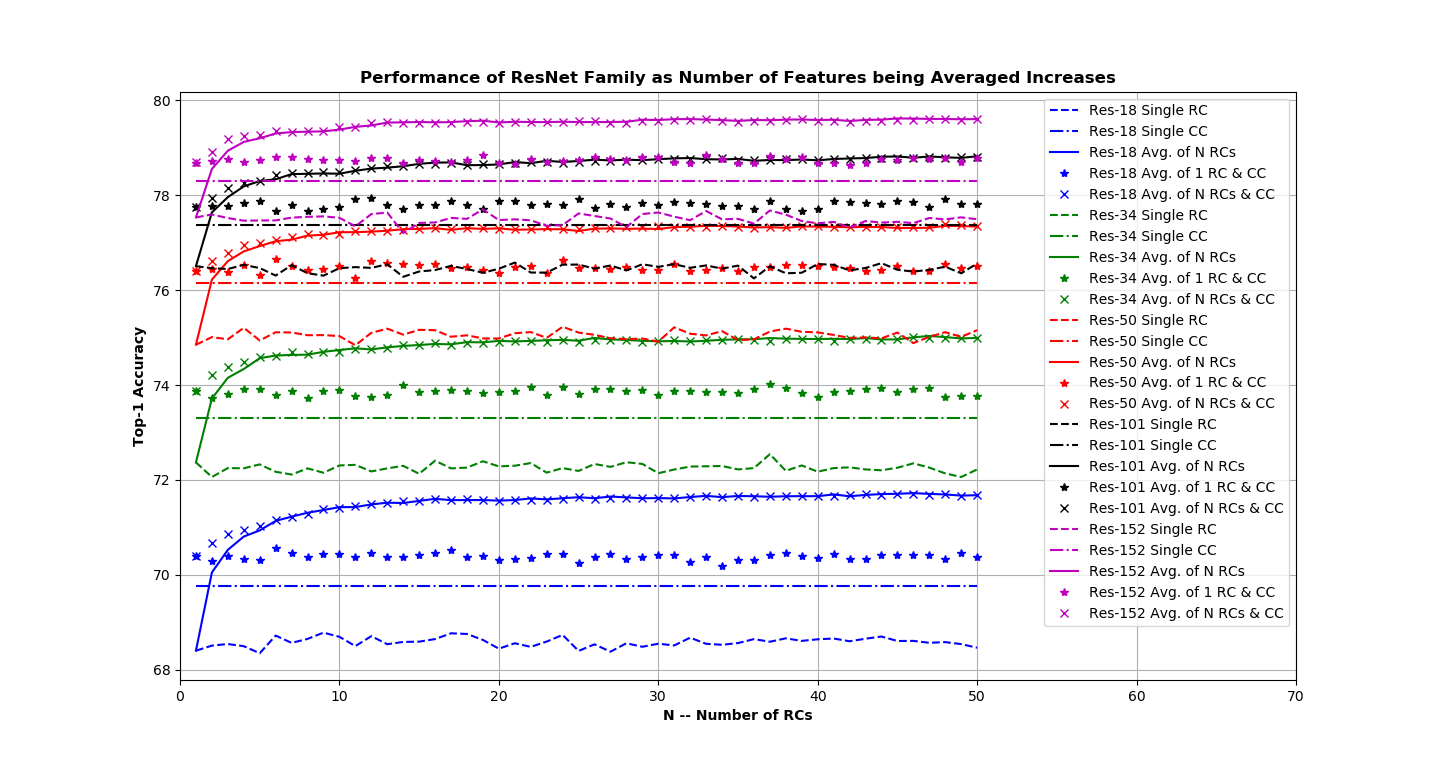}
\caption{Top-1 accuracy of ResNet family as the function of number of random crops (RCs) being averaged. We demonstrate when only a single crop is used per image, on average using central crop (CC) is better than random crop (RC). However as the number of random crops being averaged increases (even average of two random crops), the performance increases. The performance is saturated after about twenty RCs irrespective of the ResNet model. We also demonstrate that including central crop in averaging helps in improving the performance when number of random crops per image is less than five. In general using even a single random crop in addition to the central crop results in better performance than using just the central one.}
\label{fig_resnet_performance}
\end{figure*}

%-------------------------------------------------------------------------

\begin{figure}[ht!]
\centering
\includegraphics[width=0.85\linewidth]{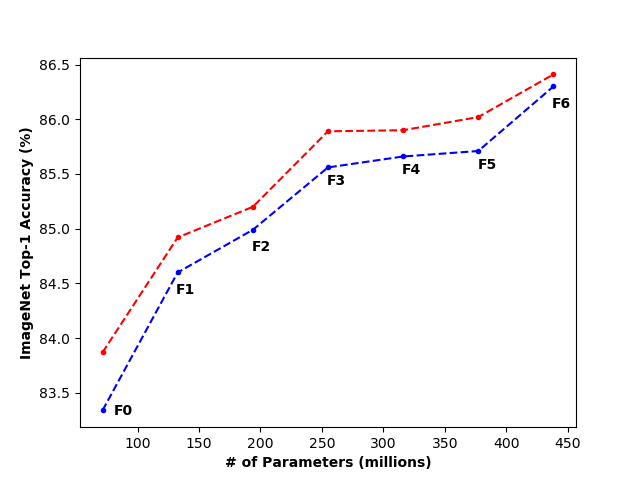}
\caption{Performance comparison of NFNet \cite{Brock2021High} compared to baselines (\textcolor{bluecol}{blue curve}) reported by Timm \cite{rw2019timm1}. Our performance (\textcolor{redcol}{red curve}) for smaller model like F3 is comparable to larger ones like F4/F5. A vendor can deploy a smaller model (fewer parameters) and still achieve performance on par with larger ones.}  
\label{fig_nfnet}
\end{figure}

\section{Discussion}
\label{sec:Discussion}

Through extensive evaluations, we have demonstrated that using central crop for inference is not optimal. Matching the train-time augmentations by using even a small number of random or fixed crops can provide performance better or comparable to approaches that require custom training and additionally leverage billion/million scale data. This is especially important for vendors from a practical deployment view point. In such practical settings, does a vendor really want to invoke a training that requires huge amounts of data and computation to train existing models just to squeeze mere $1$ -- $2$ \% extra performance? For example for NFNet \cite{Brock2021High}, a smaller model like F1 or F3 can be used to squeeze the performance equivalent to a larger model like F2 or F4/F5 (Fig \ref{fig_nfnet}). To this end, even approaches like Meal \cite{Shen2019MEAL} or Meal-V2 \cite{Shen2021MEAL} that do not rely on extra data beyond ImageNet, still require performing inference of 1.2 million training images through several teacher networks to get the softmax soft labels. A resource-constrained vendor can directly deploy the available pre-trained models with rare extra processing required for non-central or long-sided images. It is worth noting that performance gains through these inference-time augmentation are mostly beneficial for smaller networks like ResNet-18, MobileNet, and B0 etc., so additional inference does not cost much especially in modern GPU era.                 

%-------------------------------------------------------------------------
\section{Conclusions}
\label{sec:Conclusions}

We have demonstrated that using central crop at inference-time is sub-optimal and one can achieve the same performance out of pre-trained models with a minor overhead due to employing multiple crops per image at inference time. The evaluation conducted on various families of modern deep networks render the performance gains due to several custom training strategies questionable. While the experiments herein have shown that MID using just random crops improves performance, future work should examine full distributional matching including other augmentations used in training of the pre-trained networks \eg, color-jittering.  

%\input{Supp}

%\subsubsection*{Acknowledgements}
%All acknowledgments go at the end of the paper, including thanks to reviewers who gave useful comments, to colleagues who contributed to the ideas, and %to funding agencies and corporate sponsors that provided financial support. 
%To preserve the anonymity, please include acknowledgments \emph{only} in the camera-ready papers.

\bibliographystyle{ieeetr}
\bibliography{egbib}

%\clearpage
\end{document}